# Automating Construction Contract Review Using Knowledge Graph-Enhanced Large Language Models

Chunmo Zheng[1], Saika Wong[1], Xing Su[1,2], Yinqiu Tang[3], Ahsan Nawaz[1], Mohamad Kassem[4]

[1]College of Civil Engineering and Architecture, Zhejiang University, Hangzhou, China

[2]Center for Balance Architecture, Zhejiang University, Hangzhou, China

[3]PowerChina Huadong Engineering Corporation Limited, Hangzhou, China

[4]Department of Engineering, Newcastle University, Newcastle NE1 7RU, UK

ABSTRACT

An effective and efficient review of construction contracts is essential for minimizing construction projects losses, but current methods are time-consuming and error-prone. Studies using methods based on Natural Language Processing (NLP) exist, but their scope is often limited to text classification or segmented label prediction. This paper investigates whether integrating Large Language Models (LLMs) and Knowledge Graphs (KGs) can enhance the accuracy and interpretability of automated contract risk identification. A tuning-free approach is proposed that integrates LLMs with a Nested Contract Knowledge Graph (NCKG) using a Graph Retrieval-Augmented Generation (GraphRAG) framework for contract knowledge retrieval and reasoning. Tested on international EPC contracts, the method achieves more accurate risk evaluation and interpretable risk summaries than baseline models. These findings demonstrate the potential of combining LLMs and KGs for reliable reasoning in tasks that are knowledge-intensive and specialized, such as contract review.

1. Introduction

Effective contract management is crucial to the success of construction projects. Poor contract management can lead to cost overruns, schedule delays, claims and disputes, which adversely impact the project performance and the interest of all parties involved. According to the Arcadis Global Construction Dispute Report, the average value of disputes in the construction industry globally in 2023 was USD 42.8 million, with an average resolution time of 13 months. The highest value dispute reported by the respondents was 2 billion USD [1]. Chan et al. [2] identified the primary causes of these disputes as misunderstandings of contractual terms, non-compliance with contractual obligations by involved parties, errors and omissions in contract documents, and poor contract administration. Therefore, effective contract management requires a comprehensive understanding and detailed scrutiny of contracts across project lifecycle. Contract review, a core process in contract management, involves activities such as risk identification, contract inconsistency detection and contract clause interpretation [3]. Contract managers must meticulously analyze extensive contract documents, comprehend legal terminology, and navigate contractual terms to identify semantic ambiguities and unbalanced provisions within contract clauses. These tasks are typically performed manually or through handcrafted rules, both of which are time-consuming and prone to errors [4,5].

Recent years have witnessed a significant progress in applying natural language processing (NLP)-based methods for automatic contract analysis, primarily involves rule-based, machine learning-based and deep learning-based approaches. Rule-based approach predefines semantic rules to match semantic pattern in contract sentences with risk sentence patterns for risk classification and identification. Machine learning or deep learning-based models are trained on annotated contract clauses with risk labels. These models primarily serve as risk classification tools for preliminary contract reviews.

Significant efforts in this domain include automated contract requirement classification [6–8], automated contract risk identification [9,10] and missing clause detection [11,12]. These methods have facilitated various degrees of digitization and automation in managing construction contracts, laying a solid foundation for NLP-based automatic contract review systems. However, these approaches remain preliminary steps in addressing the complexity of real-world contract risk review, for the following critical gaps that have not been bridged. First, rule-based approaches rely heavily on manually crafted rules, making them time-consuming and difficult to adapt to diverse contracts and review tasks. On the other hand, machine learning-based methods require large annotated datasets, which are often expensive and labor-intensive to produce. The limitation hold back their applicability across varied contract review scenarios. Second, risk identification often demands understanding nuanced relationships within clauses and between clauses, as well as interpreting the broader implications embedded in the overall contract structure. However, state-of-the-art automated contract risk identification approaches based on machine learning or deep learning fail to capture the implicit and intricate semantics presented in the holistic context of construction contracts. Third, while significant efforts have been devoted to representing regulatory knowledge in NLP applications, the semantic representation of contract clauses remains underexplored. Current methods do not adequately model the complex semantics embedded in contract text, limiting the ability of machines to understand contract semantics at a fine-grained level. Finally, current approaches lack effective mechanisms for retrieving and synthesizing context from contracts, as well as the integration with reasoning engine for advanced contract reasoning. To address these challenges, this paper investigates how integrating KGs with LLMs can enhance the accuracy and interpretability of the automated review of construction contracts.

This paper proposes the integration of semantic-rich contract knowledge representation, efficient knowledge retrieval, and advanced reasoning capabilities using LLMs to achieve comprehensive contract risk analysis. LLMs have emerged as powerful tools for complex knowledge processing owing to their advanced language understanding and ability to handle a wide range of NLP tasks, such as question answering [13] and text summarization [14]. They are trained on massive text datasets, enabling them to answer a wide range of human queries without the need for laboriously collected, annotated domain-specific data [15]. In the construction domain, the effective utilization of LLMs has been reported as one of the most needed endeavor [16]. Researchers have also explored the use of LLMs to perform legal case analysis and contract question answering, demonstrating a great potential in using LLMs for legal processing and interpretation [17].

The black-box nature of LLMs and their lack of domain-specific knowledge often lead to hallucinations and factual inaccuracies, posing significant challenges in contract review tasks that demand precise and accurate interpretation and processing [18,19]. To address this challenge and enhance the LLM capability, recent studies have suggested the use of symbolic and explicit knowledge representations such as knowledge graphs (KGs) [20]. KGs, which capture entities and their interrelations using graph-based data model [21], explicitly express the relationships between entities and intuitively display the overall structure of knowledge, thus serving as an ideal choice for knowledge representation. Unlike existing approaches using vector similarity search to retrieve sentence-level knowledge for LLM reasoning, the KG-based knowledge retrieval allows more accurate and entity-aware knowledge enhancement.

By focusing on explicit entity relationships and structured knowledge, this approach is expected to improve both the accuracy and interpretability of LLM-assisted automated contract review. However, despite the potential of KGs, existing approaches lack a comprehensive model tailored for the automated review of construction contracts, which often involve intricate, nested logic and multi-layered semantic relationships. To address this gap, we propose the Nested Contract Knowledge Graph (NCKG), a specialized knowledge representation model designed to encapsulate the complex semantic layers inherent in construction contracts.

The NCKG comprises a nested knowledge representation layer and an ontology layer, enabling the representation of complex contract semantics for automated tasks. Furthermore, we develop and evaluate an NCKG-enhanced contract review approach that includes NCKG modeling, knowledge retrieval, and knowledge-enhanced prompting. This approach aims to enhance LLMs' capabilities in knowledge-intensive tasks, such as contract risk identification, while also providing valuable insights for other contract review tasks requiring a deep understanding of contract context.

The rest of this paper is organized as follows. Section 2 introduces the background and current challenges in automated contract review, LLMs and graph retrieval-augmented generation pattern, and contract knowledge modeling. Section 3 presents the KG-enhanced LLM approach for automated contract review and explains the involved modules. Section 4 uses risk clauses in construction projects to demonstrate our approach in comparison with two baseline methods, along with the discussion of the performance results and implications. Finally, conclusion and suggestions for further research are summarized in Section 5.

2. Literature review

*2.1 NLP solutions for automated contract review*

The utilization of Natural Language Processing (NLP) has revolutionized the analysis of voluminous texts within the construction domain. A variety of text automation tasks including information extraction (IE), text classification (TC) and question answering can be performed by NLP through lexical, syntactic and semantic-level analysis [22]. Their final goal is to enable human language to be comprehended and processed automatically by machines [23].

As many legal issues are closely associated with the use of natural language in legal documents, using NLP for reviewing and management of legal text (i.e. construction contracts) is a promising solution. Automated contract review can be categorized into rule-based, machine learning (ML)-based and deep learning (DL)-based approaches.

Rule-based approach operates on the principle of matching predefined rules, encapsulated by the "if pattern, then result" logic [24]. For instance, Al Qady and Kandil [25] developed a semantic representation framework and used shallow parser to extract elements from construction contract. Lee et al. [10] proposed a rule-based approach to match the subject-verb-object (SVO) component in each contract clause with certain risk category. If the subject belongs to "employer-side party", the verb belongs to "assign" class, and the object belongs to "right" class, then this clause is identified as assignment risk. Padhy et al. [26] developed a rule-based NLP model to identify exculpatory clauses through syntactic, lexical and semantic analysis. The rule-based approach associates the semantic concept with contract risk patterns to extract potential risk clauses, and the concept often contains legal parties, obligation, rights, legal actions and claim/dispute [3]. It has been reported that rule-based approach often yields reliable accuracy at the cost of extensive human labor involved in developing hand-crafted rules. Rule-based systems also suffer from limitation in scalability and robustness, due to the inability to operate on unfamiliar and error inputs, such as misspelled and unseen words [27].

Compared to rule-based approach, ML-based approach is considered more robust and scalable, where data patterns are learned automatically from training dataset [28]. Variety of ML algorithms containing naïve bayes (NB), support vector machines (SVM), k-nearest neighbors (KNN), logistic regression (LR), and decision tree (DT) have been tested in the automated contract review context. Hassan et al. [29] compared various supervised ML approaches including NB, SVM, LR, KNN, DT to develop an automated requirement classification framework for construction contracts. Candas and Tokdemir [30] developed NLP and ML integrated model to identify vagueness in contractual terms. From a multifunctional contract administration perspective, Yang et al. [31] developed a ML-based contract function classification model to facilitate the functional distribution of contracts. Deep learning is a rapidly growing subfield of ML, which have demonstrated its potential in NLP tasks through automatic feature extraction, facilitating the learning of intricate patterns through nonlinear parameter combinations [32]. For example, a hybrid approach combining semantic and ML techniques is developed for clause classification, achieving a recall of 100% [33]. Moon et al. [5] developed a named entity recognition (NER) model based on bidirectional long short-term memory (Bi-LSTM) architecture for construction specification review.

Despite the effectiveness and robustness of ML-based approaches, they rely heavily on high-quality annotated data for training effective models. In the construction domain, contract text annotation requires domain expertise, resulting in a scarcity of adequately annotated datasets [34]. This limitation poses significant challenges for model training, potentially restricting the practical application of these automated contract review systems. It underscores the need for more flexible and data-efficient methods for automated contract review tasks.

*2.2 Large language models and retrieval-augmented-generation (RAG) pattern*

Language models predict the next token in a sequence based on the probability distribution of preceding tokens, with deep neural network architectures serving as their backbone. Early models, such as Recurrent Neural Networks (RNNs) and Long Short-Term Memory (LSTM) networks, were widely used for processing sequential data [35]. However, these architectures faced challenges in capturing long-term dependencies and enabling parallel computation. The introduction of the Transformer architecture [36] addressed these limitations with its attention mechanism, allowing for more effective contextual representation of input text. This innovation became the foundation for state-of-the-art models such as BERT [37] and Generative Pretrained Transformers (GPT) [38], which leverage large-scale pretraining to develop general language understanding capabilities. These models can then be fine-tuned for specific tasks, including machine translation, question answering, text classification, and information extraction.

In the construction domain, the pre-trained and finetuning pattern has been adopted to solve many underlying problems such as construction knowledge retrieval [39], construction plan alignment [40], and contract management [41]. However, creating domain-specific datasets, such as those for legal or construction contexts, is resource-intensive, time-consuming, and costly, which limits the scalability of this approach for academia and industry [42]. LLMs trained on general corpora offer an alternative by enabling zero-shot reasoning, which eliminates the need for additional fine-tuning. Through prompt engineering, LLMs have demonstrated strong performance in a variety of tasks, offering simplicity, transferability, and high accuracy [43,44]. For example, Polak and Morgan [45] proposed a fully automated data extraction framework leveraging GPT-4, which required minimal manual input and demonstrated efficiency in complex scenarios.

Despite their remarkable capabilities, LLMs often exhibit "hallucinations," producing inaccurate or false outputs, particularly in domain-specific applications. This issue arises from the lack of specialized domain knowledge in their training corpora. To address this, Retrieval-Augmented Generation (RAG) has emerged as a promising approach, integrating external, expert-driven knowledge bases to enhance domain-specific reasoning capabilities [46]. RAG utilizes vector databases as external memory, converting textual data into vector representations for efficient similarity-based retrieval [47]. Wong et al. [34] demonstrated a tuning-free RAG method for automated risk identification, using a domain-specific vector database to improve the reasoning ability of LLMs. However, current RAG approaches rely on general-domain vector representations and similarity search, which often fail to capture the nuanced semantics and entity-level knowledge critical for specialized domains.

To address these challenges, Graph Retrieval-Augmented Generation (Graph RAG) has emerged as a novel approach that seamlessly integrates structured graph knowledge into the reasoning processes of LLMs [48]. By retrieving graph elements—such as nodes, triples, and subgraphs—relevant to a specific query from a pre-constructed knowledge base, GraphRAG enhances response generation with enriched contextual information. This approach leverages the interconnected nature of graph-based data, facilitating more precise and comprehensive retrieval of relational knowledge. Furthermore, knowledge graphs and similar graph-structured data provide explicit representations of entities and their interrelations, effectively complementing the implicit reasoning capabilities of LLMs.

*2.3 Knowledge modeling in construction and current challenges*

Knowledge Graphs (KGs) have gained prominence as representing knowledge in a format that is both human-friendly and machine-interpretable, which are often constructed on top of an ontology, which offer a formal mechanism for conceptualizing terms and relations within a domain [49]. In the context of the construction contract domain, researchers introduced using KGs to support automated requirement modeling for complex systems. However, most of studies did not consider the complexity in regulation and contract semantics [50]. A triple denoted as (subject, predicate, object) is the basic semantic structure of a KG which can capture the binary relations between classes. Binary relations are used to connected two entities to form a single fact such as "A is capital of B". In reality, however, a large proportion of knowledge contains multiple entities formed in a nested and complex logic. It is reported that more than 1/3 of the entities participate in non-binary relations in Freebase [51]. Those non-binary relations and constraints are important knowledge sources, which requires the extension in KG to include the metadata and n-ary relations [52].

Approaches to metadata modeling in RDF-based systems, such as standard reification, singleton property, and RDF-star, have been developed to address these limitations, representing contextual information within graph structures [53]. RDF-star, in particular, has gained attention for its ability to represent nested data structures efficiently and directly, without introducing redundant nodes [54]. These extended KG enhance the representation of complex semantics. Some attempts model the complex semantics with extended KG for facilitating the domain application. For example, Li et al. [55] extended the triples to quadruples with additional attributes in the medical KG. Yang et al. [50] proposed a knowledge graph-based modeling of design code with four semantic types—order, complex, event, and integration schemas for automatic compliance checking. Zhou et al. [56] utilized conceptual graph for interpretating rules for code compliance checking.

However, the complex knowledge representation of construction contract is not yet discovered. Contract knowledge contains intricate web of provisions and relationships, which concepts are interconnected with complex and nested relations. The semantic complexity from a standard form of construction contract compared to a single fact is depicted in Fig.1, which implies the triple structure is limited in modeling construction contract knowledge. Therefore, a formalized knowledge representation schema designed to capture the complexity nature in construction contract is necessitated.

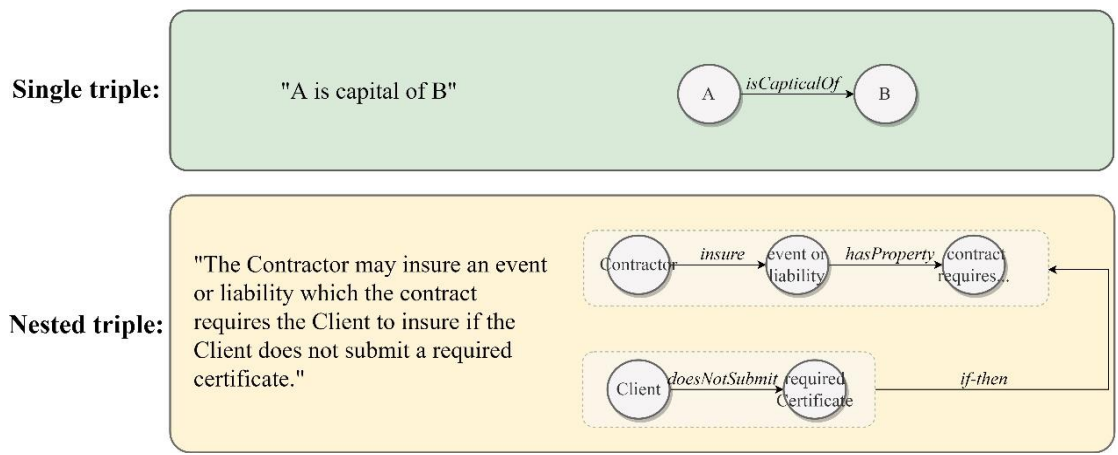

Fig. 1. Comparison of single triple and nested triple pattern

This paper aims to address the current gaps in knowledge representation within construction contracts by developing a knowledge graph modeling approach. This approach is designed to facilitate the integration of external knowledge into LLMs for enhanced understanding and processing of construction contracts. Representative studies in semantic modeling for automated contract management is shown in Table 1.

Table 1. Comparison between existing contract automation tasks enhanced with semantic modeling

| Existing approaches | Semantic modeling | | | Reasoning | | Task |
|---|---|---|---|---|---|---|
| | KR method | Enhanced with domain ontology | Complex/nested logic modeling | Reasoning model | Adaptation to KR | |
| [10] | Ontology | √ | × | Rule-based model | Pattern matching | Contract risk identification |
| [57] | Ontology | √ | × | Semantic model | Query path design | Automatic compliance checking |
| [58] | Knowledge graph | √ | √ | Deep learning-based | Query path design | Automatic compliance checking |
| [50] | Knowledge graph | × | √ | Semantic model | Query path design | Automatic compliance checking |
| [34] | Vector database | × | × | LLM | Similarity search | Contract risk identification, risk reasoning |
| Our approach | Ontology and knowledge graph | √ | √ | LLM | Natural language prompt& Query path design | Contract risk identification, risk reasoning |

3. Proposed approach: a KG-enhanced LLM method for automated contract review

The overall framework of our approach is described in this section, as illustrated in Fig. 2. The framework consists of two main components: 1) Domain NCKG development, which constructed the knowledge graph using standard form of construction contract, following LLM-enabled extraction and manual checking; 2) NCKG-enhanced contract review, which integrates graph similarity search and structured queries for knowledge retrieval, enhancing LLM-based risk analysis through knowledge-enhanced prompting. The detailed workflow of each component is elaborated below.

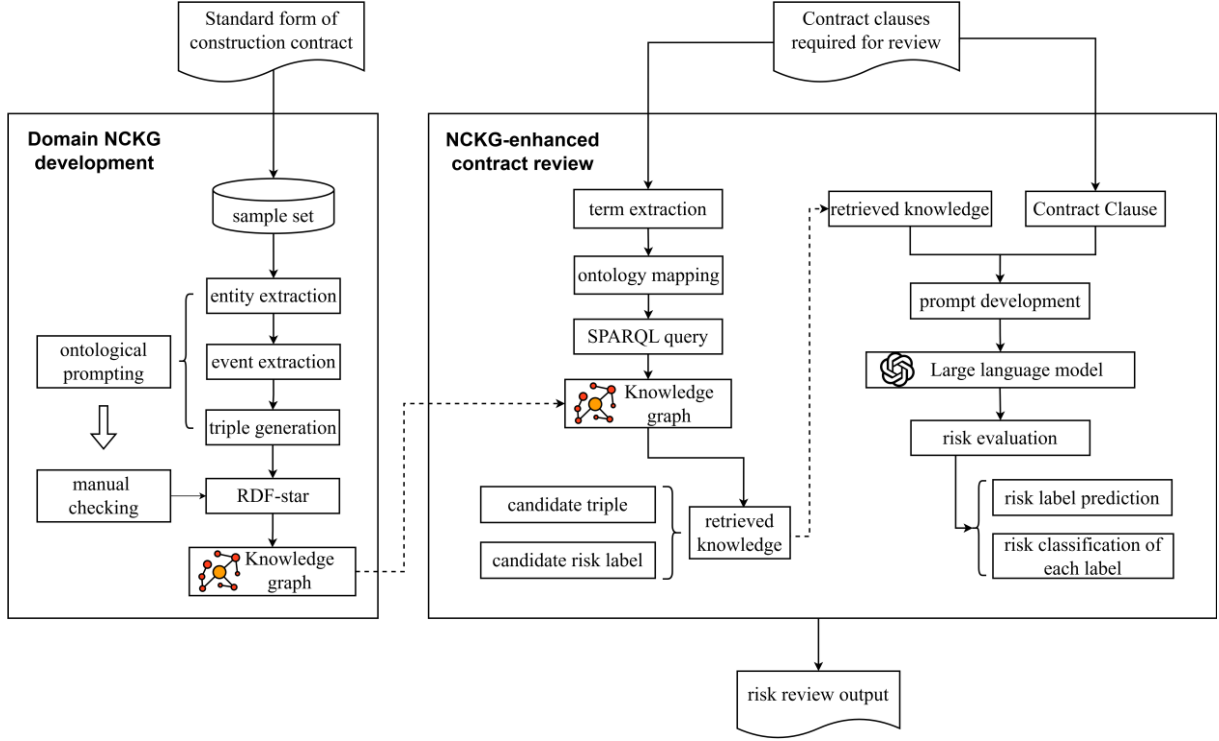

Fig. 2. Overall framework

*3.1 NCKG representation model and its modeling approach*

This section introduces the Nested contract knowledge graph (NCKG) representation model, designed to represent the intricate knowledge within construction contracts. The model comprise several integral components, including the NCKG meta model, which serves as the foundation for representing clause-level contractual knowledge, and a risk-level classification framework that integrates with the NCKG to enhance its risk reasoning capabilities. Furthermore, this section illustrates the intricate semantics representation of contract clauses, presenting methodologies for transforming these clauses into a machine-readable knowledge graph format using RDF-star. Additionally, it explores the semi-automatic approach to NCKG construction, highlighting techniques that balance automation and expert intervention to ensure both accuracy and efficiency.

Key aspects of the NCKG model are explored in this section. Section 3.1.1 presents the NCKG representation model, which defines the foundational structure and ontology used to represent clause-level knowledge. Section 3.1.2 discusses the integration of risk-level knowledge into the NCKG, enabling the classification and representation of risks associated with contract clauses. Section 3.1.3 elaborates on the representation of intricate semantics, including the use of RDF-star syntax for modeling nested relationships, and outlines the transformation process from raw contract clauses to structured knowledge graph representations. Finally, Section 3.1.4 introduces the semi-automatic construction approach for building the NCKG, leveraging LLMs with prompt-based method to streamline the extraction and representation of knowledge.

*3.1.1 NCKG representation model*

The nested knowledge representation layer is crucial for organizing and structuring data, facilitating the depiction of complex relationships within contract contexts through its utilization of triples, events, and their interconnected, nested relationships. This model extends the conventional definition of a triple in knowledge graphs by maintaining the structure of the head and tail nodes connected by a relationship edge, while allowing nodes to represent either a single entity or an event. The core components of our model include:

Entity. Represented as E, an entity is a basic unit of concept in the NCKG. It can connect with other nodes (entities or events) in a knowledge graph to form triples. Entities serve as foundational nodes that facilitate the representation of contract concepts.

Event. Represented as Evt, it is a nested node encapsulating a triple. This allows a triple to be treated as a node, enabling additional annotations or constraints. For instance, the triple <Contractor, submit, Programme> represents a statement within a contract. By framing this triple as an event node, metadata or constraints—such as submission deadlines or compliance requirements—can be attached. Event nodes, as elements of the graph, can also connect to other nodes (entities or events) to depict more complex, nested relationships, thereby enriching the semantic representation of contract contexts.

Relation. Represented as r, including the connections between nodes including entity-to-entity (E2E), entity-to-event (E2Evt), event-to-event (Evt2Evt), and compound relations (e.g. "and/or").

Triple. Represented as (h, r, t), where h denotes the head node, t denotes the tail node, and r denotes the relationship, depicted as a directed edge from the head node to the tail node. Nodes within a triple can be categorized as either an entity (node) or an event (node), with T representing a set of triples, E representing a set of entities, R representing a set of relations, and Evt representing a set of events. Our model delineates three types of triples $T_{E2E}$, $T_{E2Evt}$, $T_{Evt2Evt}$, signifying the entity-to-entity, entity-to-event and event-to-event triples, respectively (refer to Table 2). Notably $T_{Evt2E}$ is equivalent to $T_{E2Evt}$, as the head and tail nodes are interchangeable, reflective of the transition between active and passive sentence structures.

Table 2. Definition of triple and fact

| Type of Triple | Definition |
| --- | --- |
| $T_{E2E}$ | $T_{E2E} = \{(h,r,t)|h,t \in E, r \in R\}$ |
| $T_{E2Evt}$ | $T_{E2F} = \{(h,r,t)|h \oplus t \in \{E, Evt\}, r \in R\}$ |
| $T_{Evt2Evt}$ | $T_{F2F} = \{(h,r,t)|h,t \in Evt, r \in R\}$ |

By structuring knowledge into entities, triples, and nested fact events, the NCKG enables rich semantic representation and facilitates comprehensive reasoning over the intricate relationships within construction contracts. Theoretically, triples and events may be infinitely nested within our model, with the extent of nesting contingent on the sentence structure and ontological concept division, for which a universal standard is absent.

The NCKG representation model is depicted in Fig. 3, comprising the meta-structure of NCKG and the NCKG ontology. The basic component of the meta model includes the entity, event, and their inter-connected relations together to form the nested pattern of the NCKG meta model. Following the meta model, the NCKG ontology layer incorporates contract knowledge patterns to facilitate the representation of the complex semantics inherent in construction contract knowledge. This layer serves as a conceptual schema essential for the reasoning, sharing, and reuse of knowledge in future applications. The namespace for the NCKG can be referred under the prefix ckg:<http://example.org/NCKG/>.

The basic upper level of ontology classes include "ckg:ContractActor", "ckg:ContractObject", "ckg:ContractProperty" and "ckg:ContractConstraint". On the event level, the upper class is represented as "ckg:ContractEvent". Classes and subclasses are linked through hierarchy relations denoted in the figure. For instance, the "ckg:ContractActor" includes "ckg:Employer-sideParty", "ckg:Contractor-sideParty", "ckg:BothParty" and "ckg:EitherParty" subclasses. The "ckg:ContractObject" is a more inclusive and diverse class which contains concept related to construction works, liability, right and responsibility, environment and information, etc. "ckg:ContractConstraint" represent the constraints to a ContractEvent, such as "ckg:TimeConstraint", "ckg:AmountConstraint", "ckg:ConditionConstraint", etc. The "ckg:ContractProperty" represents the definition, inclusion or status of contract objects. The "ckg:ContractEvent" is the core component in NCKG, as the risks are often related to certain contractual event. The subclasses include "ckg:Instruction", "ckg:Submission", "ckg:Payment", etc.

The relations in NCKG has also entity-level and event-level relations. The entity level relation (E2E relation) connects the "ckg:ContractActor" with "ckg:ContractObject" to form the basic unit of an "ckg:ContractEvent". The E2Evt relation attached additional constraints to an event, such as "ckg:hasCondition", "ckg:hasTimeConstraint". For the Evt2Evt relation, the NCKG ontology layer defines contractual relations to represent event-level relations such as conditional and temporal relations. For example, triples like <Event1, hasCondition, Event2> or <Event1, within2weeksOf, Event2> exemplify the "hasContractualRelation" class. The first triple implies that Event1 should occur only under the condition of Event2, whereas the second indicates that Event2 should take place within two weeks after Event1. Further explanation of each class can be found in Table 3, where we provide a comprehensive explanation and illustration of each class type within the ontology layer, categorized according to the elements in the nested framework concerning entities, facts, and the relationships connecting them.

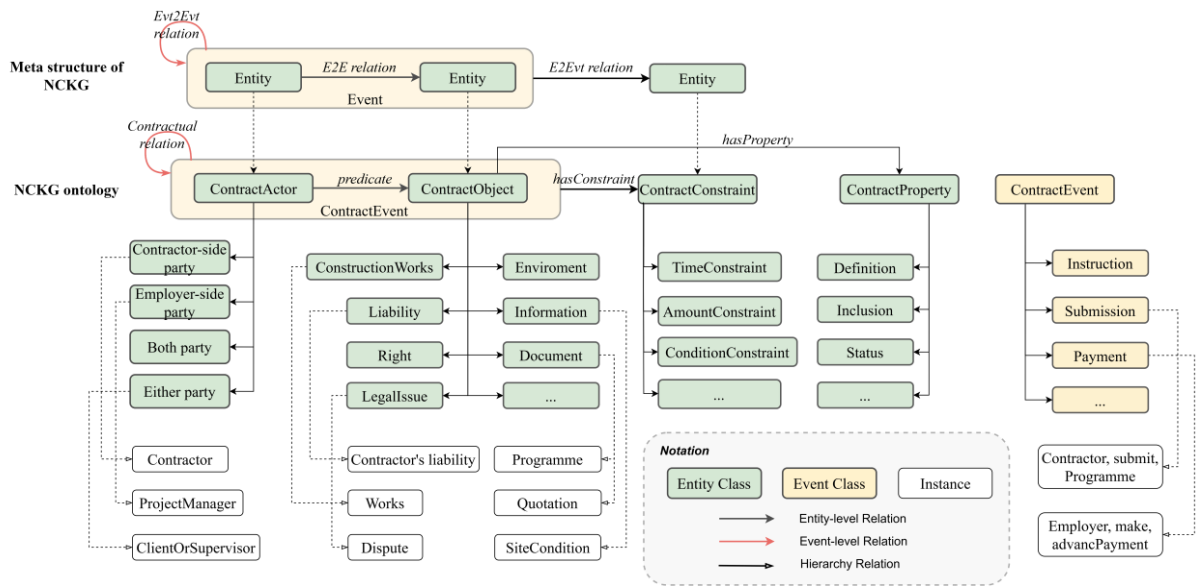

Fig. 3. NCKG knowledge representation model (partial)

**Table 3.** Definition of elements in ontology

| Type | Class name (entity/ fact/ relation) | | Explanation | Instance illustration |
|---|---|---|---|---|
| entity | ContractActor | | The parties in a contract. "Either party" denotes one of the parties is the subject of a behavior. "Both party" denotes both of the parties are the subject of a behavior. | "Client"; "ProjectManager"; "Contractor", "Subcontractor"; "Supervisor"; "ProjectManagerAndContractor", "ContractorOrSubcontractor"; … |
| | ContractObject | | The object that a ContractActor acts upon. It includes classes such as "Document", "Information", "Environment", … | "Programme"; "site information"; "early warning meeting"; "Contractor's design", "defect", "changeOfScope"; … |
| | ContractProperty | | The descriptions of the current definition, status, or inclusion, etc. of a ContractObject. | "submitted"; "isNotIdentified"; "isChanged", "isNotCompensationEvent"; … |
| | Constraint | TimeConstraint | The time or time period during which certain ContractEvent happens. | "withinOneWeekOfStartingDate; "winthinPeriod for reply"; "beforeKeyDate"; … |
| | | AmountConstraint | The required amount, currency of certain ContractActor conducting a certain ContractEvent. | "in the amount and currencies stated in the Contract Data"; … |
| | | ResultConstraint | The expected result of certain ContractActor conducting a certain ContractEvent. | If an event occurs which stops the Contractor from completing the whole of the works. "the failure of completing whole of the works" is a Result_constraint. |
| | | ConditionConstraint | The precondition of certain ContractActor conducting a certain ContractEvent. | The Contractor may publicise the works only with the Client's agreement. "Client's agreement" is a ConditionConstraint |
| relation | E2E relation | hasProperty | The relation that connects a ContractObject and a ContractProperty entity. | "hasProperty"; "hasInclusion"; "hasDefinition"; … |
| | | predicate | The relation used for linking ContractActor and ContractObject in a ContractEvent. | "submit"; "instruct"; "makePayment"; … |
| | E2Evt relation | hasConstraint | The relation that connects ContractEvent and Constraints. | "hasTimeConstraint"; "hasAmountConstraint"; "hasResultConstraint"; "hasConditionConstraint"; One of the example triples: <Event, hasTimeConstraint, beforeReplyDueDate>. |
| | | | | The Event could be <ProjectManger, agreeTo, extension>. |
| | Evt2Evt relation | Conditional relation | The conditional relation between two Events. It is connected by conditional identifiers such as "ckg:hasCondition" "ckg:IfNot-then" or "ckg:unless". | "ckg:hasCondition" (notice that <A, hasCondition, B> denotes "if B, then A".); "ifNot-then", "otherwise" (<A, ckg:ifNot-then, B> or <A, ckg:otherwise, B> denotes "if not A, then B".); "unless" (<A, ckg:unless, B> denotes if B, then not A) |
| | | Temporal relation | The temporal relation between two Events to form the triple <$Event_1$, hasTimeConstraint, $Event_2$>. The relation should be replaced by the real conditional relation such as "within2weeksOf". | The temporal relations including "before"; "after", "until", "within 2 weeks of", "as soon as", etc. Notice that <A, before, B> denotes that A happens before B happens |
| event | | payment | The event of the Employer making payment to the Contractor. | <Employer, make, advancePayment> |
| | | issurance | The event of the Employer or the Engineer formally issuing a certificate, notice, or other document as required under the contract. | <ProjectManager, issue, certificate>; |
| | | instruction | The event of the Employer or the Engineer giving an instruction to the Contractor regarding the execution of the works. | <Employer, giveInstruction, startOfWork> |
| | | submission | The event of the Contractor submitting documents, plans, or other deliverables required by the contract to the Employer or the Engineer for review or approval. | <Contractor, submit, Programme> |

3.1.2 *Risk-level knowledge integration*

The risk-level knowledge representation aims to link the clause-level knowledge with the high level risk catgories. This paper classifies contractual risk into seven categories based on literature [59,60]. The risk categories include "Assignment", "Payment", "Temporal", "Financial", "Differing site condition (DSC)", and "Liability". These labels are used to identify risks in contract clauses, and one or more categories may be associated with a single clause requiring review. The definitions of the seven risk labels are detailed in Table 4. Additionally, three risk types are introduced to indicate the specific risk associated with each risk category:

Ambiguity: Refers to unclear or vague language in the clause that could lead to misinterpretation or disputes.

Unbalanced Obligation: Refers to clauses that disproportionately favor one party, creating inequitable responsibilities.

No Risk: Refers to clauses that are clear, balanced, and present no significant risk.

Table 4. Definition of risk categories

| Risk category | Definition |
| --- | --- |
| Assignment | Risks related to the transfer of rights, benefits, or obligations under the Contract by either party. |
| Payment | Risks associated with the timing, adequacy, or guarantee of payments from the Employer to the Contractor. |
| Temporal | Risks linked to the project timeline, including delays, extensions, and scheduling uncertainties. |
| Financial | Risks involving financial stability, funding guarantees, or changes in the financial conditions of the Employer. |
| DSC | Risks arising from differing site conditions, such as unforeseen physical conditions or subsurface variations. |
| Liability | Risks tied to the allocation and limitation of liabilities, including financial exposure for damages or claims. |

Fig. 4 illustrates the mapping from contract instances to the NCKG ontology layer and subsequently linking them to the associated risk categories. The risk labels are connected to the NCKG ontology classes through the "ckg:hasRiskCategory" relationship. Contractual risks often arise from specific contractual events or constraints. Consequently, the ontology integrates classes for events and constraints, which are mapped to their respective risk categories, enabling a systematic risk identification process. For example, in the clause "Employer shall pay to Contractor within 90 days of the submission of Payment application", the following entities and events are identified that form the basis for linking clause-level knowledge to the ontology layer. The identified entities include "Employer", "Contractor", and the "Payment Application", each representing essential components of the contractual relationship. These entities are interconnected through two events: "ckg:Employer ckg:makePaymentTo ckg:Contractor", which denotes the Employer's obligation to make payment to the Contractor, and "ckg:Contractor ckg:submit ckg:PaymentApplication", which represents the Contractor's submission of the Payment Application. Additionally, a time constraint is identified in the relationship between the two events, i.e., the Employer must make payment to the Contractor within 90 days of the submission of the Payment Application. These instances are then linked to their respective upper-level ontology classes via the "rdf:type" relationship. For example, the event

"ckg:makePaymentTo" is linked to the ontology class "ckg:Payment" and subsequently assigned the risk label "Payment" The time constraint identified within the events is linked to "ckg:TimeConstraint", which corresponds to the risk label "Temporal".

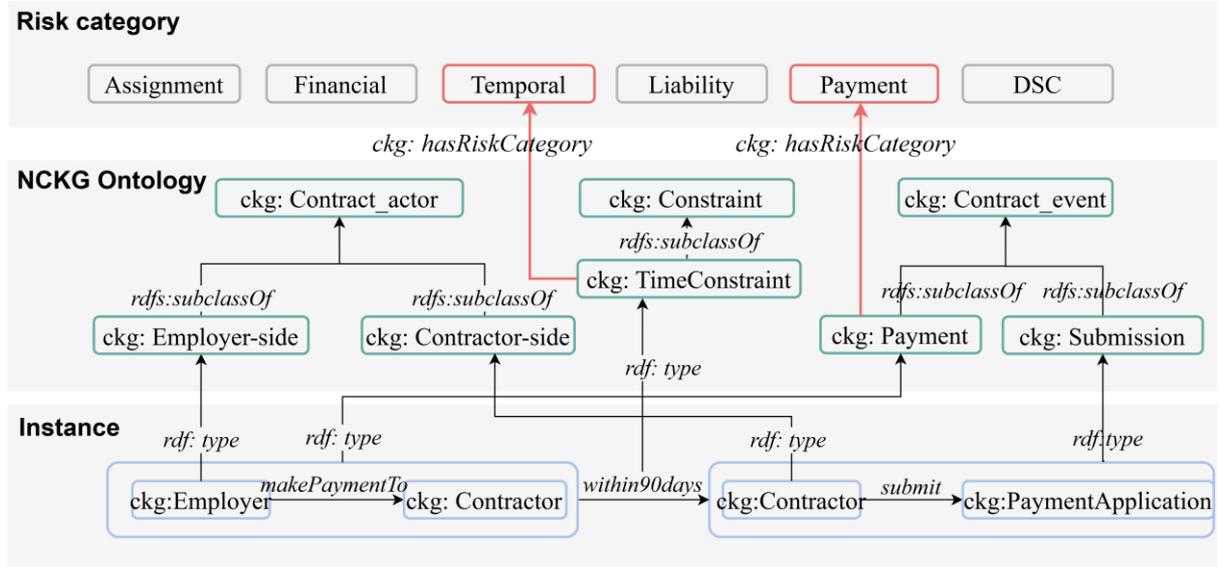

Fig. 4. Knowledge mapping from contract instance to risk category

The integration of clause-level knowledge with risk-level knowledge allows the mapping from contract clauses to their respective risk category during the contract knowledge retrieval process. Furthermore, the risk categories serve as pre-identifications for large language model (LLM)-based risk reasoning by providing candidate labels for machines to analyze specific risks. Compared to traditional machine learning methods, such as training a classifier to identify risks in clauses, the proposed approach leverages a knowledge graph to link risk categories to entity- and event-level representations. This integration provides explicit relationships between contractual events, constraints, and their associated risks, which enhances the interpretability in risk reasoning process.

*3.1.3 Intricate semantics representation and clause-to-rdf transformation*

The representation of nested relationships in NCKG leverages RDF-star syntax, which extends the RDF framework to annotate triples and represent nested relations. This extension allows for the modeling of not only simple entity-to-entity relationships but also more complex relationships such as conditional, causal, and coordinating relations, effectively capturing the intricate semantics of contract clauses.

As illustrated in Table 5, these complex relationships are defined using specialized predicates in the ontology, such as "ckg:hasCondition", "ckg:exception", and "ckg:hasInclusion", to represent conditional relations, exceptions, and compound relationships, respectively. These predicates are primarily used to indicate relationships between events, enabling a nuanced understanding of contractual semantics.

It is worth noting that even the same type of relationship can have varied representations depending on contractual practices. For instance, in the clause "The Contractor and the Client provide materials, facilities, and samples for tests and inspections as stated in the Scope", the word "and" appears three times, but each instance is represented differently.

1. "The Contractor and the Client": This involves two instances of the ContractActor class, where both parties share the responsibility for providing resources. This relationship is represented using two separate triples.

2. "Materials, facilities, and samples": These represent multiple inclusions of the resources to be provided. They are connected to the resource entity using the ckg:hasInclusion predicate to indicate that these items collectively form the required resources.
3. "Tests and inspections": This phrase refers to a construction process that consistently appears as a single conceptual unit. It is represented as one instance, TestAndInspection, rather than multiple entities.

Table 5. NCKG representation of complex relations

| Complex type | Relation in NCKG | clause example | RDF-star representation |
|---|---|---|---|
| Conditional relation | ckg:hasCondition | The Contractor may insure an event or liability if the Client does not submit a required certificate. | PREFIX ckg:<http://example.org/NCKG/> <<ckg:Contractor ckg:insure ckg:eventOrLiability>> ckg:hasCondition <<ckg:Client ckg:doesNotSubmit ckg:requiredCertificate>> . |
| Conditional&Exception relation | ckg:hasCondition ckg:exception | 72.1 The Contractor removes Equipment from the Site when it is no longer needed unless the Project Manager allows it to be left in the works. | PREFIX ckg:<http://example.org/NCKG/> <<ckg:Contractor ckg:removeFromSite ckg:Equipment>> ckg:hasCondition <<ckg:Equipment ckg:hasProperty ckg:notNeeded>> ; ckg:exception <<ckg:ProjectManager ckg:allows ckg:EquipmentLeftInWorks>> . |
| Coordinating relation (Compound subject, compound object) | ckg:hasInclusion | The Contractor and the Client provide materials, facilities and samples for tests and inspections as stated in the Scope. | PREFIX ckg: <http://example.org/NCKG/> <<ckg:Contractor ckg:provide ckg:Resources>> ckg:hasPurpose ckg:TestAndInspection . <<ckg:Client ckg:provide ckg:Resources>> ckg:hasPurpose ckg:TestAndInspection . |

ckg:resource ckg: hasInclusion ckg: materials;

ckg: hasInclusion ckg: facilities;

ckg: hasInclusion ckg: samples .

To illustrate the transformation process from contract clauses to RDF-star representation, an example is provided in Fig. 5. This example demonstrates how a clause from the NEC contract is transformed into an RDF-star representation, effectively capturing its intricate semantics. The clause includes multiple events and complex relationships, such as temporal relations (e.g., actions constrained by timeframes) and coordinating relations (e.g., multiple statements interconnected within a single clause). These relations showcase the capability of the proposed NCKG approach to handle complex scenarios where multiple relationships coexist.

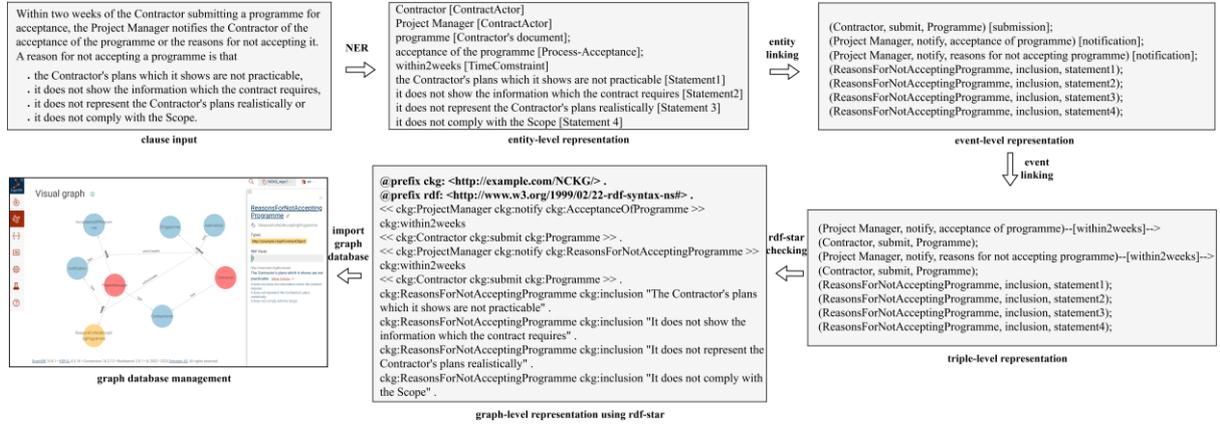

Fig. 5. Clause-to-rdf transformation process for NCKG construction

### 3.1.4 Semi-automatic construction of NCKG

The transformation task can be defined as taken a contract clause input C, convert it to NCKG = {Ent, R, Evt, Const}, where Ent = {e1, e2, …, em} denoting a collection of m different entities in the knowledge graph; R = {r1, r2,…, rk} denoting a collection of k different relations between entities; Evt = {evt1, evt2, …, evtn} representing n different event, each of event evt = (ei, r, ej) is a triple formed by two entities connected by a relation; Const = {c1, c2, …, cn} representing constraints attached to the event node, differentiated from the one connected to an entity. The NCKG construction process is to extract the component in NCKG sequentially, converting the original clause C to entity-level, event-level, triple-level and then graph-level representation. To automate this process, traditional NLP techniques using dependency parsing suffers from handling complex multi-level nested relations, leading to incomplete parsing trees. This method leverages LLMs to extract multi-layer contract knowledge using a prompting engineering approach.

The extraction algorithm is shown in Fig. 6. Taken a contract clause C and the NCKG ontology as input, the algorithm first extract the fundamental event knowledge in the clause by performing named entity recognition of "ContractActor" and "ContractObject". The ContractEvent are then formed in the triple format of (ContractActor, predicate, ContractObject). The next steps extract the ContractProperty and forms the (ContractObject, hasProperty, ContractProperty) triple. The ContractConstraint entity is then extracted. With all the entities and events recognized, the final relation linking step extract nested triples by linking event&constraint and event& event. The nested triples are (evt, hasConstraint, ContractConstraint) and (evt, hasContractualRelation, evt).

The main functions for NCKG modeling include named entity recognition and relation linking, which are implemented using prompt engineering approaches. Detailed prompts are provided in Fig. 7.

| ALGORITHM 1: NCKG MODELLING |
|---|
| **Input:** $C \leftarrow$ contract clause, $NCKG\_onto=\{ContractActor, ContractObject, ContractEvent, Constraint, Property\}.class \leftarrow$ NCKG ontology |
| **Initialize:** $Ent \leftarrow[], R \leftarrow[], Evt \leftarrow[], Triple \leftarrow[]$ |

| | |
|---|---|
| 1 | **for** *clause in C* |
| 2 | **do** |
| 3 |    *ContractActor, ContractObject $\leftarrow$ NamedEntityRecognition(NCKG_onto.ContractActor), NamedEntityRecognition(NCKG_onto.ContractObject)* |
| 4 |    **if** *ContractActor and ContractObject* **then** |
| 5 |       *(ContractActor, predicate, ContractObject) $\leftarrow$ RelationLinking(ContractActor, ContractObject)* |
| 6 |       *R.append(predicate); Ent.append(ContractActor, ContractObject, ContractProperty)* |
| 7 |       **if** *predicate* **then** *Evt.append(SetEvent(ContractActor, predicate, ContractObject))* |
| 8 |    **end** |
| 9 |    *ContractProperty $\leftarrow$ NamedEntityRecognition(NCKG_onto.ContractProperty)* |
| 10 |    **if** *ContractObject and ContractProperty* **then** |
| 11 |       *(ContractObject, hasProperty, ContractProperty) $\leftarrow$ RelationLinking(ContractObject, ContractProperty)* |
| 12 |       *Evt.append(SetEvent(ContractObject, 'hasProperty', ContractProperty))* |
| 13 |       *Ent.append(ContractProperty)* |
| 14 |    **end** |
| 15 |    *ContractConstraint $\leftarrow$ NamedEntityRecognition( NCKG_onto.ContractConstraint)* |
| 16 |    **for** *evt, constr in zip(Evt, ContractConstraint or [])* **do** |
| 17 |       **If** *constr* **then** *Triple.append(SetTriple[(evt, 'hasConstraint', constr), (evt, 'hasContractualRelation', evt)])* |
| 18 |    **end** |
| 19 | **end** |

Fig.6. NCKG instance extraction algorithm

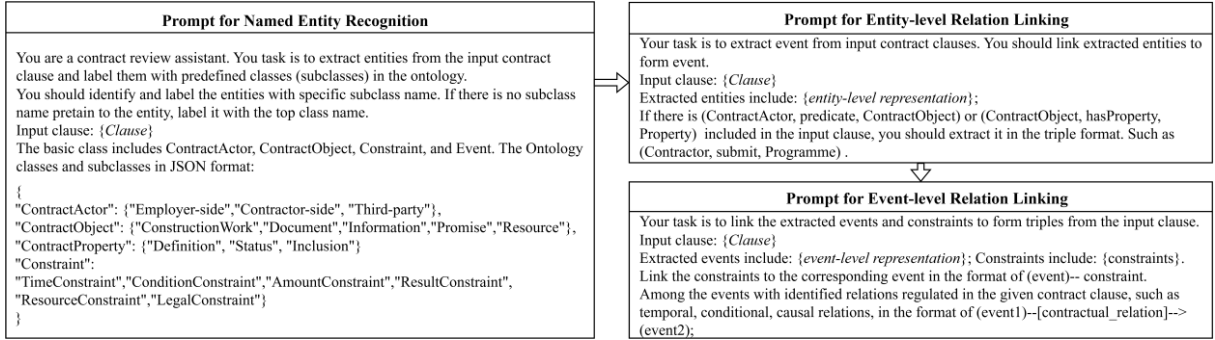

Fig.7. Named entity recognition and relation linking prompt

*3.2 Knowledge graph-enhanced approach for automated contract review*

To facilitate precise and entity-level interconnected knowledge representation for contract risk identification and review, we propose the NCKG-enhanced automated contract review approach, as illustrated in Fig. 8. This method follows an RAG paradigm, comprising two primary stages: knowledge retrieval and knowledge-enhanced prompting.

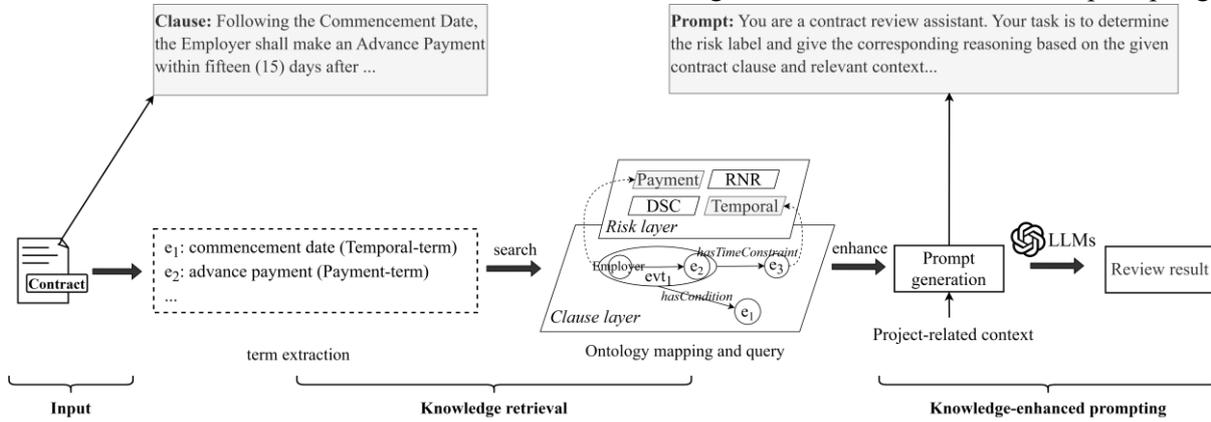

Fig. 8. NCKG-enhanced LLMs for automated contract review approach

3.2.1 Knowledge retrieval

The knowledge retrieval process aims to extract, map, and retrieve contextual knowledge relevant to key contractual terms identified in an input contract clause. This stage consists of three main steps: (i) term extraction, (ii) ontology mapping, and (iii) knowledge retrieval in the NCKG database. The detailed procedure is outlined in Algorithm 2 and illustrated in Fig. 9.

Term extraction involves performing named entity recognition (NER) on the input contract clause to identify key contractual terms. This extraction is facilitated by an LLM, which is prompted to output two key concepts that encapsulate the core contractual topics. The model is instructed to prioritize terms reflecting critical contractual aspects, including Assignment, DSC, Financial, Payment, Temporal, and Liability, while avoiding generic contract terms such as party names or contract prices. The following structured prompt is employed:

"You are a construction contract assistant. Extract two key concepts or entities from the following contract clause, representing the main topics discussed. Avoid generic terms such as party names and contract prices. Contract Clause: {clause}. Output your response as a comma-separated list of two entities."

For instance, in the example contract clause shown in Fig. 8, the extracted terms "commencement date" and "advance payment" are identified as the most representative contractual concepts.

Once the key terms are extracted, they are mapped to semantically relevant entities or events in the NCKG ontology using similarity search. A TF-IDF-based similarity score is computed between the extracted terms and NCKG elements to determine the best matches. The similarity function is defined as:

$$TF - IDF_{i,j} = tf_{i,j} \times log\left(\frac{N}{dfi}\right)$$

where $tf_{i,j}$ represents the term frequency of entity $i$ in document $j$, $dfi$ is the document frequency, and $N$ is the total number of documents. Candidate entities and events with the highest similarity scores are stored in entity_map and event_map, respectively. For example, the extracted term "commencement date" matches entity "ckg:commencement" and "ckg:commencementDate", and the term "advance payment" matches entity "ckg:advance payment" and the event <<ckg:Employer ckg:make ckg:advancePayment>>.

The mapped entities and events serve as query anchors to retrieve contractual context from the NCKG database via SPARQL queries. This retrieval consists of two key steps: clause-level context retrieval and risk label prediction. In the clause-level context retrieval step, The system executes SPARQL queries to retrieve both single and nested triples containing the identified entities and events. These triples contains the candidate entities and events, forming the contract context. The retrieved context may include contractual events, liabilities, and constraints associated with the identified terms, offering a comprehensive understanding of the clause. For example, the nested triple "ckg:commencement ckg:hasCondition <<ckg:Employer ckg:make ckg:advancePayment>>" in NCKG contains with both the matched entity "ckg:commencement" and "ckg:advance payment", it indicates the commencement should condition on the advance payment made by the Employer. The second step is risk-level context search, which identifies the candidate risk label associated with the retrieved triples from the last step. The associated risk labels are also searched through the SPARQL query in the NCKG. These risk labels are critical for downstream tasks, as they provide explicit aspects for risk prediction and analysis.

The output of the whole knowledge retrieval stage is the "retrieved_triple" and "retrieved_risk_category", which are subsequently integrated into the "contract_review_prompt" for the LLM-based automated contract review.

3.2.2. Knowledge-enhanced prompting for automated contract review

The prompt template, as shown in Fig. 10, is designed to standardize the process of contract risk identification and analysis. Its key components include the input contract clause, relevant context providing additional background on contractual terms or related events, and a candidate risk category label. LLMs are instructed to perform two tasks: risk evaluation and risk summary.

In risk evaluation, the LLM systematically assesses whether the clause presents a risk under each risk category. The identified risk types include Ambiguity, referring to unclear language that may lead to disputes, and Unbalanced Obligation, indicating a clause that disproportionately favors one party. Each clause is assigned a risk classification in the format [Risk Category]-[Risk Type] (e.g., Payment-No Risk), ensuring structured and interpretable outputs.

For the risk summary, the model generates a concise synthesis of the clause's overall risk status. This summary provides a brief yet comprehensive explanation, enabling stakeholders to quickly grasp the nature and severity of potential contractual risks.

By structuring the prompt in this way, the contract review system ensures a domain-specific, context-aware, and interpretable evaluation of contractual risks, enhancing the reliability of risk assessments in construction contracts.

| | ALGORITHM 2: NCKG RETRIEVAL |
|---|---|
| | **Input:** $C \leftarrow$ contract clause |
| | **Initialize:** $extracted\_term \leftarrow []$, $entity\_map \leftarrow []$, $event\_map \leftarrow []$, $retrieved\_triple \leftarrow []$, $retrieved\_risk\_category \leftarrow []$ |
| | Set up openAI, pygraphdb API |
| 1 | **for** *clause* in $C$ |
| 2 | **do** |
| 3 |    $term \leftarrow$ NamedEntityRecognition(*clause*) |
| 4 |    *extracted_term*.append(*term*) |
| 6 |    $entity_i$, $event_i \leftarrow$ GraphSimilaritySearch(*i*); |
| 7 |    *mapping_to_entity*.append($entity_i$); *mapping_to_event*.append($event_i$) |
| 8 |    **if** $entity_i$ **and** $event_i$ **then** |
| 9 |       $triple \leftarrow$ sparqlQuery(""" SELECT ?s ?p ?o WHERE { { ?s ?p $entity_i$ } UNION { $entity_i$ ?p ?o } } """) |
| 10 |       $nested\_triple \leftarrow$ sparqlQuery(""" SELECT ?s ?p ?o WHERE { { ?s ?p $event_i$ } UNION { $event_i$ ?p ?o } } """) |
| 11 |       *retrieved_triple*.append([*triple, nested_triple*]) |
| 12 |       $r \leftarrow$ sparqlQuery(""" SELECT ?r WHERE { triple hasRiskCategory ?r } """) |
| 13 |       *retrieved_risk_category*.append(*r*) |
| 14 |    **end** |
| 15 |    **return** *retrieved_triple, retrieved_risk_category* |
| 16 |    *review_output* $\leftarrow$ openai.chat(*clause, contract_review_prompt, retrieved_triple, retrieved_risk_category*) |
| 17 |    **return** *review_output* |
| 18 | **end** |

Fig. 9. NCKG-enhanced RAG algorithm

| **Prompt for contract risk identification and analysis** |
|---|
| You are a construction contract review assistant. Your task is to determine the risk category of the input clause and give the corresponding reasoning based on the given contract clause and relevant context. You will be given a) the input contract clause; b) the related context related to terms or events in the clause; c) the potential risk category in the contract.<br>Input contract clause: {*input_clause*};<br>Relevant context: {*retrieved_triple*};<br>Risk category:{*retrieved_risk_category*};<br>Your task is to:<br>1) Risk evaluation:<br>For each risk label, identify whether the clause presents the following risk type:<br>    Ambiguity: Unclear language or interpretation that may lead to disputes.<br>    Unbalanced Obligation: A clause that disproportionately favors one party.<br>    No Risk: Clear and balanced terms with no significant issues. If the clause is clearly and balanced provision, you should honestly reply "No risk".<br>Return the conclusion in the format of "[Risk category]-[Risk type]". e.g. Payment-No risk;<br>2) Risk Summary:<br>Based on your evaluation, provide a concise risk review summary (max 100 words) synthesizing the overall risk status of the clause. |

Fig. 10. Contract risk identification and analysis prompt

## 4 Experimental results

The effectiveness of the proposed approach is evaluated on risk clauses covering various risk categories and is compared against baseline methods in both the knowledge retrieval and risk review stages. This section details the datasets, gold standards, baseline methods, and evaluation metrics used in the experiments.

*4.1 Data preparation*

Data preparation involves constructing external knowledge databases from standard form of construction contract and collecting test data from international construction project contracts. The NCKG database was constructed in this paper following the semi-automated approach described in Section 3. A total of 143 clauses from FIDIC and NEC are used to construct the NCKG for testing. The sections in FIDIC incorporate sections including "General", "Employer's Financial Arrangements", "Commencement of Works", "Time for Completion", "Advance Payment". And the NEC sections contain identical sections including "General Provisions", "The Contractor's main responsibilities". "Time", "Payment". The source of the clauses and the number of extracted triples is shown in Table 6. A total of 335 triples were extracted for the NCKG database, and 179 of them are nested triples which have additional constraints attached to the contractual event. These contract data are stored and visualized in GraphDB as shown in Fig. 11, enabling further retrieval and interaction with reasoning engine.

Table 6. NCKG database composition

| Data source | No. of clause stored in NCKG | No. of triple extracted | No. of nested triple extracted |
|---|---|---|---|
| FIDIC red/silver book | 57 | 154 | 89 |
| NEC | 86 | 181 | 90 |
| Total | 143 | 335 | 179 |

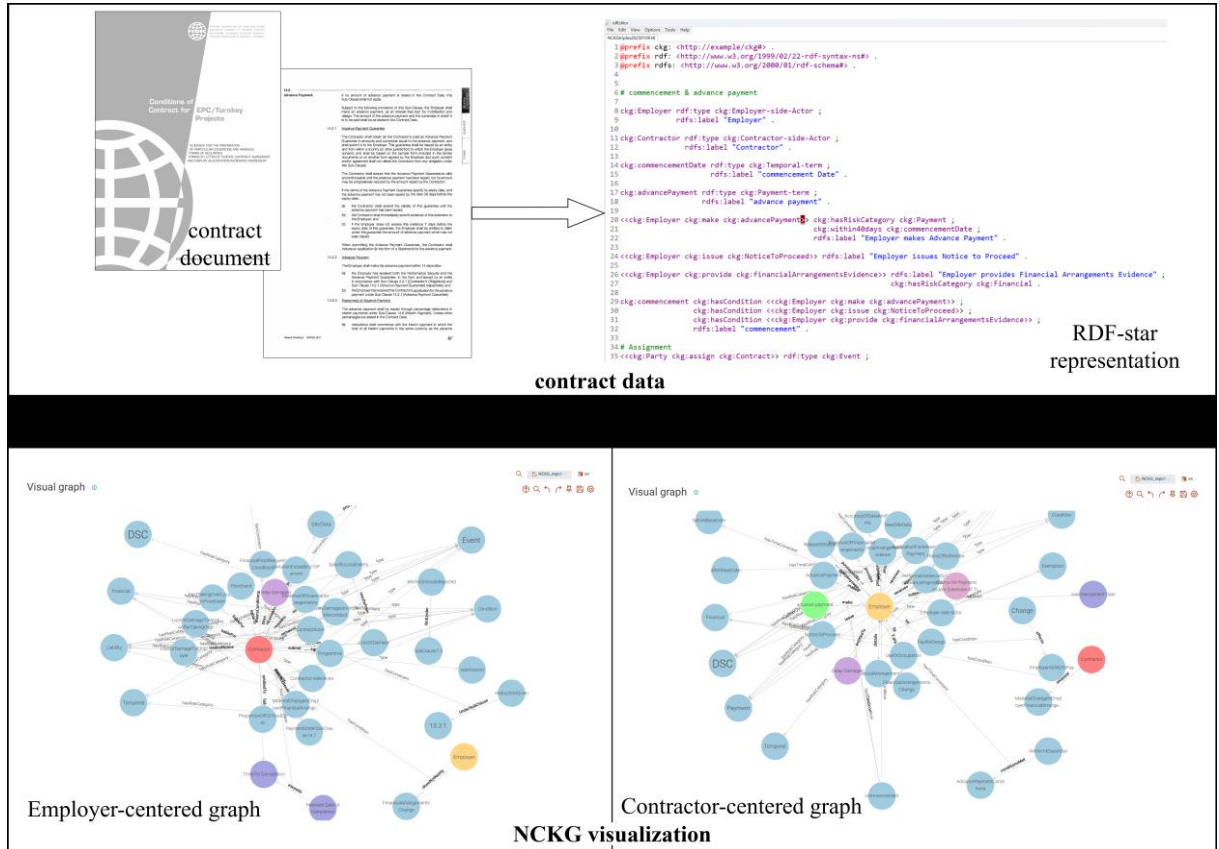

Fig. 11. Established KG in GraphDB based on contract data

Contract risk clauses were collected from international construction projects and utilized as test data in the case study. A total of 18 risk clauses were annotated with predefined risk categories, associated risk types, and expert-provided risk summaries for each clause. These clauses encompass all six risk categories and were drafted using FIDIC and NEC templates, as well as non-standard formats, reflecting the diversity of contract types in the test data. To establish a reliable ground truth for risk identification, a roundtable discussion was conducted with five domain experts, whose backgrounds are detailed in Table 7. Initially, each expert independently reviewed the case study and assigned two-level risk labels to the clauses. Following this, the experts engaged in a discussion to reconcile their findings and reach a consensus on the risk evaluations. An example of the annotated dataset is presented in Table 8.

Table 7. Description of expert background

| Expert ID | Field of Expertise | Years of Experience | Number of Previous Projects Reviewed | Current Position |
| --- | --- | --- | --- | --- |
| Expert 1 | Construction Law and Contract Management | 3 | appr. 25 | Legal Consultant, Construction Firm |
| Expert 2 | Risk and Claim Management in Construction Projects | 8 | appr. 100 | Senior Contract Expert, Construction Firm |
| Expert 3 | Project Management and Dispute Resolution | 5 | appr. 60 | Contract Manager, Construction Firm |

| Expert 4 | Project Management and Investment | 12 | appr. 40 | Department Manager, Construction Firm |
| Expert 5 | Project Development and Management | 15 | appr. 30 | Freelance Consultant |

Table 8. Example from annotated risk clause dataset

| Case number | Contract clause | Risk category | Risk type | Risk summary |
|---|---|---|---|---|
| 1 | The Employer shall pay to the Contractor:(a) the down-payment in the amount of [10%] of the Contract Price within 40 days of the Commencement Date. | Payment; Financial | Unbalanced obligation | This clause presents a Payment and Financial Risk to the Contractor, as it may require starting work before receiving the down-payment. If the Employer delays payment, the Contractor could cover upfront costs for mobilization and initial work, which increases financial pressure. |
| 2 | The Project Manager certifies a payment within a reasonable time period following each assessment date. | Payment | Ambiguity | The term "within a reasonable time period" leaves room for subjective interpretation. This vagueness can lead to disagreements between the parties as to what constitutes a "reasonable" timeframe for certifying payments. |
| 3 | It shall be a condition precedent to the issuance of the Notice to Proceed that each of the following has occurred, unless any such condition precedent is waived in writing by Project Company in its sole discretion. | Liability | Unbalanced obligation | The Employer's right to unilaterally waive certain conditions could result in ambiguities or disputes later if the waived conditions lead to unforeseen issues. This may expose the Contractor to liability for delays or defects stemming from circumstances that the conditions precedent were meant to mitigate. |
| 4 | If the Contractor encounters unforeseen physical conditions materially different from those in the Contract, they shall promptly notify the Engineer. Upon confirmation, the Contractor may be entitled to an | DSC | No risk | The "Unforeseen Physical Conditions" clause poses no risk as it ensures balanced handling of differing site conditions. It requires timely Contractor notification and provides fair entitlement to time extensions and cost |

| extension of time and reimbursement of additional costs, subject to reasonable mitigation efforts. | adjustments, subject to the Engineer's confirmation and mitigation efforts. |
|---|---|

*4.2 Baselines and evaluation method*

Two baseline methods including "LLM-only" and "vector database-enhanced" scenario were adopted to demonstrate the effectiveness of our NCKG-enhanced LLMs for automated contract review, and the detailed implementation and prompt is illustrated below. The details of each baseline method are explained as below.

**LLM-only scenario**. The LLM-only baseline uses LLMs to evaluate and analyze the risk in the input contract clauses. The prompting process is without the retrieval process from any external database, while other parts of the prompt remains exactly same with our approach. The input clause is concatenated with the standard prompt template for prompting LLM and derived a risk review output. The standard prompt template is:

"You are a construction contract review assistant. You will be provided with a) the input contract clause; b) the risk labels.

Input contract clause: {clause};

Risk label: "Assignment", "Payment", "Temporal", "Financial", "DSC", "Liability";

Your tasks are:

1) Risk evaluation: determine one or more correct risk labels. For each risk label, identify whether the clause presents the following risk type:

Ambiguity: Unclear language or interpretation that may lead to disputes.

Unbalanced Obligation: A clause that disproportionately favors one party.

No Risk: Clear and balanced terms with no significant issues. If the clause is clear and balanced provision, you should honestly reply "No risk".

2) Risk Summary: Based on your evaluation, provide a concise risk review summary (max 100 words) synthesizing the overall risk status of the clause."

**Vector database-enhanced scenario.** This baseline method takes the input contract clause and use the contract clause to retrieve similar clause from vector database. The vector database stores identical knowledge from the standard form of construction contract described in data preparation section using the Milvus vector database. This baseline method is designed to examine the difference between sentence-level similarity search and the entity-level knowledge graph search in the knowledge retrieval process, and how they affect the contract review result.

retrieval is implemented through vector similarity search, and the similarity score is derived using cosine similarity, as described below.

$$cosine\ similarity = \cos(\theta) = \frac{A \cdot B}{\|A\|\|B\|} = \frac{\sum_{i=1}^{n} A_i B_i}{\sqrt{\sum_{i=1}^{n} A_i^2} \cdot \sqrt{\sum_{i=1}^{n} B_i^2}} \quad (1)$$

In the vector database-enhanced scenario, the contract clause with the highest similarity score is retrieved as external knowledge, concatenated with the original prompt, and then fed back into the LLMs to generate a review output. The vector database-enhanced prompt template is:

"You are a construction contract review assistant. You will be provided with a) the input contract clause; b) the standard provision in FIDIC; c) the risk label.

Input contract clause: {clause};

Standard provision: {standard_provision};

Risk label: "Assignment", "Payment", "Temporal", "Financial", "DSC", "Liability";

Your tasks are:

1) Risk evaluation: for each risk label, identify whether the clause presents:

Ambiguity: Unclear language or interpretation that may lead to disputes.

Unbalanced Obligation: A clause that disproportionately favors one party.

No Risk: Clear and balanced terms with no significant issues. If the clause is clear and balanced provision, you should honestly reply "No risk".

Answer in the format: [Risk label]--[Risk type] such as [Assignment]-[No risk].

2) Risk Summary: Based on your evaluation, provide a concise risk review summary (max 100 words) synthesizing the overall risk status of the clause."

**Evaluation metrics**. To evaluate the outcomes of the three case studies, we developed an evaluation metric comprising two dimensions: the Risk Evaluation score (RE-score) and the Risk Summary score (RS-score). The RE-score assesses the accuracy of risk category classification by comparing the predicted labels with the ground truth labels. The evaluation adopts a multi-label classification framework based on a confusion matrix, categorizing predictions for each label into true positive (TP), false positive (FP), true negative (TN), and false negative (FN). For example, for the "Assignment" label: TPs are clauses correctly identified as having an "Assignment" risk. FPs are clauses wrongly predicted to have an "Assignment" risk when they do not. TNs are clauses correctly identified as not having an "Assignment" risk. FNs are clauses that actually have an "Assignment" risk but were not identified as such. The F1-score is used as the primary performance metric, computed by taking the macro-average across all classification labels. The F1-score is calculated as follows.

$$Precision = \frac{TP}{TP+FP} \quad (1)$$

$$Recall = \frac{TP}{TP+FN} \quad (2)$$

$$F1 = \frac{2 \times Precision \times Recall}{Precision+Recall} \quad (3)$$

By using macro-averaging, the evaluation treats all labels equally, regardless of their frequency, ensuring that the model performs well across all risk categories.

The RS-score, on the other hand, evaluates the quality and relevance of the risk summaries generated for each clause. This score is determined by comparing the model-generated risk summaries with expert-annotated summaries, based on metrics such as semantic similarity, coverage of key information, and clarity. Together, the RE-score and RS-score provide a comprehensive evaluation of the model's ability to accurately classify risks and generate meaningful summaries for contract clauses.

*4.3 Evaluation and case illustration*

The case study utilizes two advanced language models: GPT-4 from OpenAI and LLaMA-3 from Meta. The performance results of our method against two baseline methods using various models are shown in Table 9.

Table 9. Performance of LLMs on risk review task

| Model | LLM-only | | Vector database-enhanced | | NCKG-enhanced (ours) | |
| --- | --- | --- | --- | --- | --- | --- |
| | RE-score(%) | RS-score(%) | RE-score(%) | RS-score(%) | RE-score(%) | RS-score(%) |
| GPT-4o | 58.7 | 78.9 | 58.7 | 80.8 | 71.1 | 84.4 |
| LLaMA3 | 57.9 | 77.2 | 57.9 | 76.7 | 64.3 | 81.4 |

The performance results, evaluated based on the accuracy of predicted labels (RE-score) and expert evaluations of review analysis quality (RS-score), highlight the effectiveness of the proposed NCKG-enhanced approach across both GPT-4 and LLaMA-3 models. When applied to GPT-4, the method achieves a RE-score of 71.1% and an RS-score of 84.4%, significantly outperforming the baseline LLM-only and vector database-enhanced approaches. Similarly, for LLaMA-3, the NCKG-enhanced approach yields a RE-score of 64.3% and an RS-score of 81.4%, demonstrating consistent improvements in label prediction accuracy and summary quality. These results confirm the robustness of the NCKG-enhanced approach in reducing hallucinations, enhancing contextual understanding, and delivering coherent, high-quality outputs, underscoring its value for automated contract risk review.

A sample case illustration is provided to demonstrate our method through the whole RAG process. The input clause in the sample case study is "The Employer shall pay to the Contractor:(a) the advance payment in the amount of 10% of the Contract Price within 40 days of the Commencement Date." This clause poses "Payment" and "Financial" risk to the Contractor as it stipulates the commencement should be prior to the receival of advance payment. In the knowledge retrieval process as shown in Fig. 12, two terms in the clause "advance payment" and "commencement" are extracted as the retrieval entities, which are used to match the entity and event in the NCKG database. Through semantic similarity search, one entity "advance payment" is returned as the matched entity with a similarity score of 1, and another event <<Employer, make, advancePayement>> is also retrieved as related event with a similarity score of 0.93. For the term "commencement", two entities including "commencement" and "commencement date" are retrieved.

The top 2 entities and events for both "extracted_term_1" and "extracted_term_2" are used for matching relevant triples in the database as context knowledge. As shown in the relevant context search box, three nested triples are retrieved. For example, "ckg:commencement ckg:hasCondition <<ckg:Employer ckg:make ckg:advancePayment>>" indicates that the Contractor's commencement should be condition on the advance payment of the Employer. At the risk category search step, relevant risk category "Payment" and "Financial" are retrieved as the candidate risk label for LLM risk review. The retrieval process in the graph database presents visualized results, while in our case studies, the results are implemented and analyzed using Python.

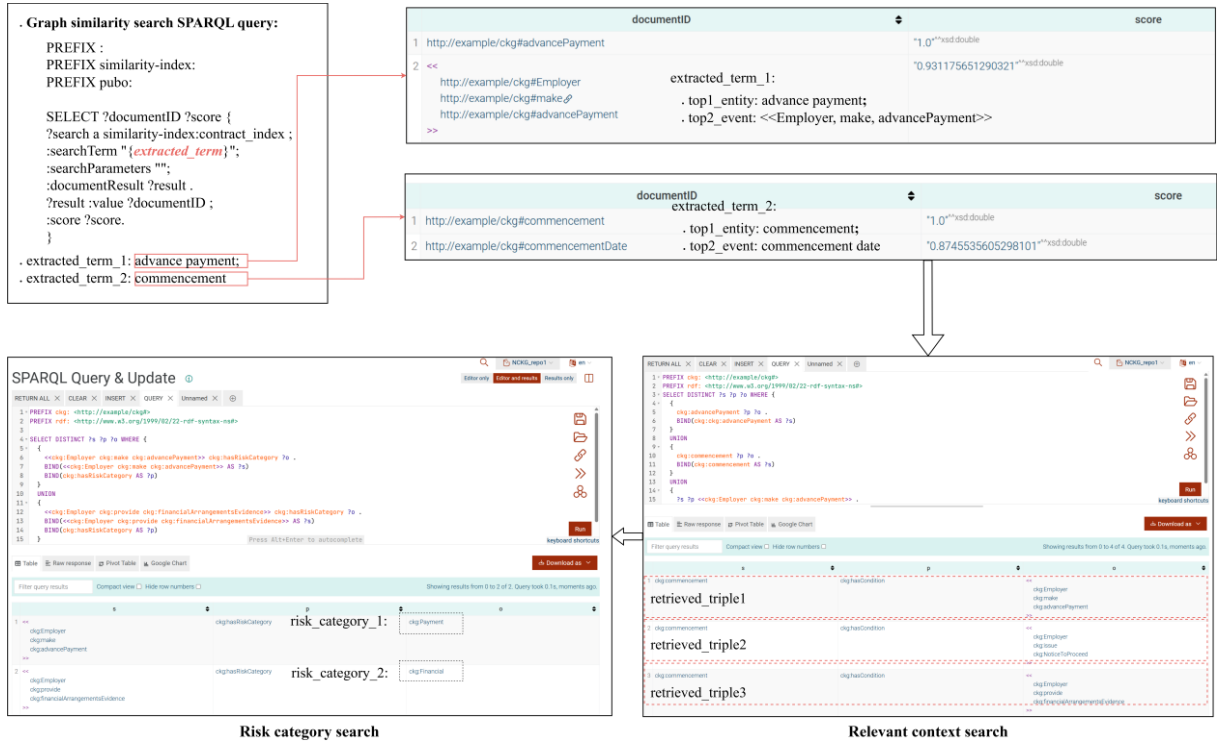

Fig. 12. Knowledge retrieval illustration in sample case study

The retrieved context and risk category are embedded into the risk review prompt and serve as input for the NCKG-enhanced contract review approach. The outputs of our approach, along with two baseline methods utilizing GPT-4o and LLaMA3, are presented in Table 10. Based on the retrieval results, clause retrieval from the SFCC using vector similarity search returns a clause from the FIDIC Red Book related to advance payment guarantees and the commencement of works section; however, it does not provide relevant context regarding the potential risks involved. In contrast, our method retrieves conditions for the commencement of works, highlighting potential risks if the payment and the employer's financial arrangements are not prepared before the commencement of works.

From the risk evaluation results, the standard approach outputs "No risk" for all labels, whereas the vector database-enhanced and NCKG-enhanced approaches successfully identify risks under the "Payment" and "Financial" categories. Furthermore, the risk summaries generated by the standard and vector database-enhanced approaches exhibit higher levels of hallucination compared to our method. Our approach, in contrast, demonstrates a more coherent and expert-driven analysis, outperforming the baselines in both precision and relevance.

Table 10. Full output on sample case study

| | | Standard | VectorDB-retrieved | NCKG-retrieved |
|---|---|---|---|---|
| Retrieved knowledge | | -- | The Employer shall deliver the guarantee to the Contractor within 28 days after both Parties have entered into the Contract Agreement. The guarantee shall be in the form annexed to these Special Provisions, or in another form acceptable to the Contractor. Unless and until the Contractor receives the guarantee, the Engineer shall not give the notice under Sub-Clause 8.1 [Commencement of Works]. | ckg:commencement ckg:hasCondition <<ckg:Employer ckg:make ckg:advancePayment>> ; ckg:hasCondition <<ckg:Employer ckg:issue ckg:NoticeToProceed>> ; ckg:hasCondition <<ckg:Employer ckg:provide ckg:financialArrangementsEvidence >> . <<ckg:Employer ckg:provide ckg:financialArrangementsEvidence >> ckg:hasRiskLabel ckg:Financial . <<ckg:Employer ckg:make ckg:advancePayment>> ckg:hasRiskLabel ckg:Payment |
| Model-1 (GPT-4o) | Risk evaluation | Payment--No risk Temporal--No risk Financial--No risk | Assignment--No risk DSC--No risk Financial--Unbalanced Obligation Payment--Unbalanced Obligation Temporal--Unbalanced Obligation | Payment-- Unbalanced Obligation Financial--Unbalanced Obligation |
| | Risk summary | The clause clearly outlines the payment terms, specifying a down-payment of 10% of the Contract Price to be made within 40 days of the | The input contract clause presents an unbalanced obligation primarily in terms of financial and payment aspects, as it lacks a payment | As the conditions for commencement, such as advance payment, notice to proceed, and evidence of financial arrangements, |

| | | | | | |
|---|---|---|---|---|---|
| | | Commencement Date. The terms are straightforward and balanced, with no ambiguity or unbalanced obligations identified. The clause effectively addresses the financial, payment, and temporal aspects without presenting any significant risks. | guarantee from the Employer, unlike the standard FIDIC provision. Additionally, the temporal aspect is unbalanced due to the extended 40-day payment period compared to the standard 28-day requirement for providing a payment guarantee. These factors could potentially favor the Employer and pose risks to the Contractor. | are not explicitly defined, leading to potential ambiguity and disputes to Contractor. It is advisable to clarify these terms to ensure balanced and clear contractual obligations. |
| Model-2 (LLaMA3) | Risk evaluation | Payment--No risk<br><br>Financial--No risk<br><br>Temporal--No risk | Assignment--No risk<br><br>DSC--No risk<br><br>Financial--Unbalanced Obligation<br><br>Payment--Unbalanced Obligation<br><br>Temporal--Ambiguity<br><br>Variation--No risk<br><br>Liability--Unbalanced Obligation | Payment--Ambiguity;<br><br>Temporal--Unbalanced Obligation;<br><br>Financial--Ambiguity |
| | Risk summary | The clause presents no significant risks, with clear and balanced terms regarding payment, financial obligations, and timelines. The language is unambiguous, and the obligations are proportionate, favoring neither party disproportionately. Overall, the risk status of this clause is low. | The clause presents some risks, primarily related to financial and payment obligations. The Employer's payment obligations are clearly stated, but the consequences of non-payment or late payment are not specified, creating an unbalanced obligation. Additionally, the temporal aspects of the clause may lead to ambiguity if the deadline is missed. | The clause specifies a down-payment amount and timeframe, but the conditions for the Commencement Date and the implications of not meeting the payment deadline are not detailed here, potentially leading to disputes. |

## 5 Discussion

This research develops a knowledge representation model to represent the intricate semantics in construction contract, and a knowledge-augmented generation approach leveraging NCKG to enhance automated contract review via LLMs. The proposed approach includes NCKG development, knowledge retrieval, and knowledge-enhanced prompting, with detailed prompts designed for both database construction and reasoning tasks.

The contributions of this paper are as follows. First, it introduces a structured, semantic-rich approach capable of capturing the nuanced and nested relationships inherent in construction contracts. By representing complex provisions such as conditional, temporal, and coordinating relationships, the NCKG addresses a critical gap in existing contract analysis methods, enabling a more accurate and contextual understanding of contract semantics. Second, the integration of NCKG with Graph RAG enhances the reasoning capabilities of LLMs. This approach allows LLMs to retrieve entity- and event-level contexts, enabling more precise risk classification and sophisticated contract reasoning, moving beyond traditional text categorization to support complex, interdependent semantic interpretations. Third, this research demonstrates the potential of combining symbolic knowledge representations, as exemplified by the NCKG, with the generative and analytical strengths of LLMs to automate domain-specific tasks in construction contract management. By offering a scalable and effective solution for knowledge representation, knowledge retrieval, and risk analysis, this paper contributes significantly to advancing automated contract review and lays the groundwork for more intelligent contract management systems in the construction industry.

Unlike existing rule-based or machine learning classifiers, which are limited to surface-level risk categorization [61,62], this method addresses the intricate semantics and interdependencies inherent in construction contracts. It demonstrates how formalized schemas, such as NCKG, enable LLMs to reason more effectively about domain-specific knowledge, overcoming the common limitations of black-box models.

The paper suggests that to accurately represent and utilize construction contract knowledge, particularly in digital and automated environments, a formalized schema capable of accommodating nuanced and nested relationships is essential. The paper demonstrated how the NCKG enables LLMs to access and interpret contract knowledge with greater precision. This advancement addresses a critical challenge in LLM research, where models often struggle with understanding and reasoning about complex, domain-specific knowledge. The NCKG's ability to represent intricate and nested relationships within contract provisions enhances the LLM's capability to process and generate accurate and contextually appropriate outputs.

The paper instigates domain researchers to further investigate more complex and hierarchical knowledge representation models tailored to the specific needs of construction management specifically and the construction industry more broadly. It also invites domain researchers to adapt existing ontologies or develop new ones to provide the foundation for KGs and other representation schemas that can work alongside LLM or other ML techniques. By pursuing these avenues, domain research can build upon the foundation laid by this paper to further advance the understanding and application of knowledge representation and retrieval in construction contract management and beyond. Notably, this research provides a basis for developing an intelligent system user interface for contract review, facilitating more efficient and informed decision-making in contract analysis.

The main limitations of this research are identified as follows: 1) While a semi-automatic method is introduced to reduce manual effort in knowledge graph construction, the NCKG method still requires resource-intensive steps for manual validation and checks, potentially limiting its scalability. Future research should aim to balance the cost of domain database development with the method's complexity, exploring more automated validation techniques and streamlined construction processes. 2) The contract review process is inherently dynamic, with contracts often subject to amendments, new clauses, and evolving legal environments. However, the current knowledge representation model does not include mechanisms for dynamically updating the knowledge graph to accommodate these changes. Future research should focus on developing automated update mechanisms, such as testing and integrating knowledge extraction algorithms or workflows to seamlessly incorporate new clauses. These limitations also highlight the complementary relationship between LLMs and KGs. LLM-enhanced KGs leverage the language generation capabilities of LLMs to help automate KG construction, while KG-enhanced LLMs incorporate structured and explicit knowledge to support domain-specific reasoning. For instance, techniques such as Chain-of-Thought and Graph-of-Thought prompting equip LLMs with symbolic reasoning capabilities, enabling more advanced and context-aware contract analysis. Future research should investigate and compare these approaches in the context of knowledge-intensive reasoning for contract review. 3) Although the current evaluation method using RE-score and RS-score accounts for label correctness and expert-rated risk analysis, more objective evaluation metrics are needed to assess LLM-generated text, reduce hallucination, and create a more reliable automatic contract review system. Future work could explore techniques such as semantic entropy evaluation [63] to enhance reliability. 4) While this study considers multiple standard forms of construction contracts, further testing is needed to validate the approach's generalizability across diverse contract types and clause structures. Construction contracts can vary significantly across industries, regions, and project requirements, and the ability to adapt the method to these variations is essential. Future efforts should focus on extending the method to include diverse contract formats, such as procurement agreements or service contracts, to ensure robustness and scalability. 5) The current approach does not fully incorporate project-specific context, such as schedules, budgets, and technical specifications, nor does it integrate expert rationales for interpreting complex or ambiguous clauses. These elements are critical for addressing nuanced and interdependent relationships within construction contracts. Future work should aim to include project context data and expert insights, potentially through hybrid approaches combining symbolic reasoning, contextual data integration, and human-in-the-loop mechanisms to enhance decision-making in complex scenarios. 6) While this study evaluates a selected set of LLMs, it does not cover all emerging commercial and open-source models, such as DeepSeek R1. As LLM capabilities continue to advance and costs decrease, integrating a broader range of models into our approach becomes increasingly feasible. Future research should explore comparative evaluations of diverse LLM architectures to assess their effectiveness in contract review tasks, ensuring a more comprehensive understanding of their strengths and limitations within this domain.

## 6    Conclusions

This study introduces a KG-enhanced LLM approach for automated contract review, leveraging the Nested Contract Knowledge Graph (NCKG) to model the intricate semantics of construction contracts. Furthermore, a NCKG-enhanced contract review method,that includes NCKG modeling, knowledge retrieval, and knowledge-enhanced prompting, was developed and tested. Experimental results demonstrate its superiority over LLM-only and vector database-enhanced methods in accuracy, interpretability, and scalability. These findings underscore the efficacy of combining explicit domain knowledge with LLMs to address challenges in contract review, particularly in the context of construction contract management.

However, realizing the full potential of this method requires several advancements. Future efforts should focus on developing fully automated or hybrid approaches to streamline NCKG construction and validation, reducing the reliance on manual processes while maintaining accuracy. Mechanisms for real-time updates to the knowledge graph are also critical, enabling the approach to adapt to evolving clauses, amendments, and regulatory changes. Expanding testing across diverse contract types and clause structures will ensure the approach's generalizability and robustness in varied contexts, while integrating project-specific contextual data and expert rationales, will enhance its ability to interpret complex interdependencies and ambiguous clauses. Additionally, improving evaluation methodologies, such as adopting advanced techniques to assess semantic consistency and reduce hallucination, will increase the method's reliability and effectiveness. These enhancements will not only address current challenges but also position the method as a dynamic and scalable solution for intelligent contract management.

In conclusion, this study advances automated contract review by combining the structured reasoning capabilities of knowledge graphs with LLMs. By addressing the outlined challenges and pursuing these enhancements, the method holds promise as a practical and scalable approach to improving contract management, contributing to progress in both research and real-world applications.

DATA AVAILABILITY STATEMENT:

All data, models, and code generated or used during the study appear in the submitted article.

ACKNOWLEDGEMENT

This research was funded by the National Natural Science Foundation of China (grant no. 71971196), Center for Balance Architecture, Zhejiang University and China Postdoctoral Science Foundation (No. 2023M743307). The authors would like to thank the 2024 European Conference on Computing in Construction (EC3) for recognizing this work as one of the best papers and for inviting the manuscript for extension and potential publication in this journal.

This research involved five domain experts who independently reviewed and annotated contract risk clauses before participating in a roundtable discussion to achieve consensus. The process focused exclusively on professional judgments, with no personal or sensitive information collected. All experts provided informed consent prior to their participation. Data collected from respondents was used exclusively for research purposes. The identities of the respondents will remain confidential and comply with all prevailing ethical regulations of Zhejiang University.